\documentclass[11pt,a4paper]{article}
\usepackage[hyperref]{emnlp-ijcnlp-2019}
\usepackage{times}
\usepackage{latexsym}
\usepackage{graphicx}
\usepackage{multirow}
\usepackage{booktabs}
\usepackage{url}
\usepackage{paralist}
\usepackage{comment}
\usepackage{etoolbox}
\usepackage{xspace}
\usepackage{color}
\usepackage[normalem]{ulem}
\usepackage[font=small,skip=0pt]{caption}
\usepackage{adjustbox}
\usepackage{dirtytalk}
\usepackage{epigraph}

\usepackage{url}

\graphicspath{ {.} }

\aclfinalcopy 

\belowdisplayskip \abovedisplayskip

\newlength{\sectionReduceTop}
\newlength{\sectionReduceBot}
\newlength{\subsectionReduceTop}
\newlength{\subsectionReduceBot}
\newlength{\abstractReduceTop}
\newlength{\abstractReduceBot}
\newlength{\captionReduceTop}
\newlength{\captionReduceBot}
\newlength{\listReduceTop}
\newlength{\listReduceBot}
\newlength{\subsubsectionReduceTop}
\newlength{\subsubsectionReduceBot}

\newlength{\eqnReduceTop}
\newlength{\eqnReduceBot}

\newlength{\horSkip}
\newlength{\verSkip}

\newlength{\figureHeight}
\newcommand{\xhdr}[1]{\vspace{3pt}\noindent\textbf{#1}}

\definecolor{redcol}{rgb}{1, 0, 0}

\newcommand{\gordonremove}[1]{
}

\title{Sunny and Dark Outside?! Improving Answer Consistency\\ in VQA through Entailed Question Generation}

\author{\textsuperscript{1}Arijit Ray, \hspace{1pt}\textsuperscript{1}Karan Sikka, 
  \hspace{1pt} \textsuperscript{1}Ajay Divakaran, 
  \hspace{1pt} \textsuperscript{*}Stefan Lee, 
  \hspace{1pt} \textsuperscript{1}Giedrius Burachas \\
\textsuperscript{1}{\tt first.last@sri.com}, \textsuperscript{*}{\tt steflee@gatech.edu}\\
\textsuperscript{1}SRI International, \textsuperscript{*}Georgia Institute of Technology
  }

\date{}

\begin{document}
\maketitle
\begin{abstract}
 While models for Visual Question Answering (VQA) have steadily improved over the years, interacting with one quickly reveals that these models lack consistency.
 For instance, if a model answers ``red'' to ``What color is the balloon?'', it might answer ``no'' if asked, ``Is the balloon red?''. These responses violate simple notions of entailment and raise questions about how effectively VQA models ground language. 
In this work, we introduce a dataset, ConVQA, and metrics that enable quantitative evaluation of consistency in VQA. For a given observable fact in an image (e.g.~the balloon's color), we generate a set of logically consistent question-answer (QA) pairs (e.g.~Is the balloon red?) and also collect a human-annotated set 
of common-sense based consistent QA pairs (e.g.~Is the balloon the same color as tomato sauce?). 
Further, we propose a consistency-improving data augmentation module, a Consistency Teacher Module (CTM). CTM automatically generates entailed (or similar-intent) questions for a source QA pair and fine-tunes the VQA model if the VQA's answer to the entailed question is consistent with the source QA pair. We demonstrate that our CTM-based training improves the consistency of VQA models on the ConVQA datasets and is a strong baseline for further research.
\end{abstract}

\section{Introduction}
\label{sec:introduction}

\setlength{\epigraphwidth}{0.925\columnwidth}
\renewcommand{\epigraphflush}{center}
\renewcommand{\textflush}{flushepinormal}
\renewcommand{\epigraphsize}{\footnotesize}
\epigraph{``A skeptic, I would ask for consistency first of all.''}
{\textit{Sylvia Plath \cite{plath2007unabridged}}}
\vspace{-5pt}

\begin{figure}[t]
\centering
\includegraphics[height=82px]{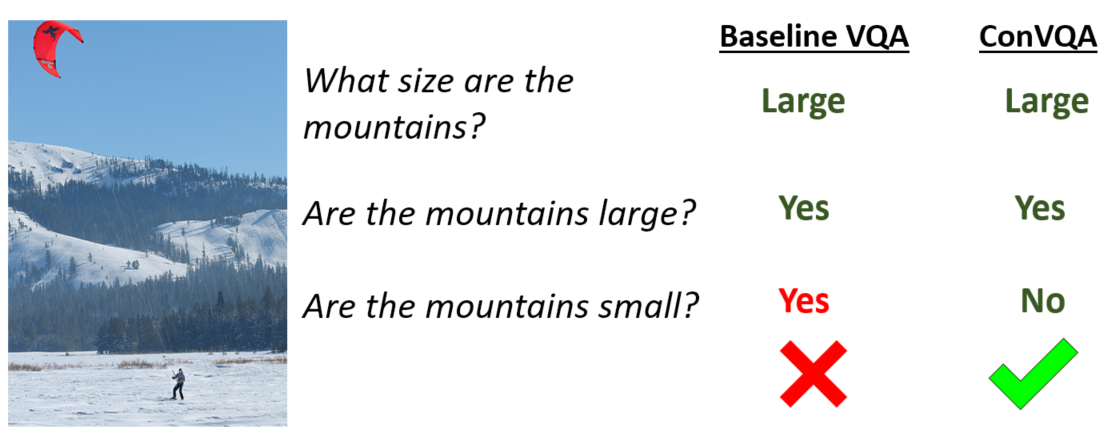}
\vspace{2mm}
\caption{Current VQA models often fail at consistently answering semantically rephrased questions. To address this limitation, we construct a consistent VQA (ConVQA) dataset with diverse QA pairs that query the same visual fact. We also propose a Consistency Teacher Module (CTM) that improves VQA consistency by rewarding consistent behavior.}
\label{fig:teaserfig}
\vspace{-2mm}
\end{figure}

\noindent Visual Question Answering (VQA) \cite{antol2015vqa} involves answering natural language questions about images. 
Despite the recent progress on VQA,  
we observe that existing methods are prone to making blatant mistakes while answering questions regarding the same
visual fact but from slightly different perspectives (Figure \ref{fig:teaserfig}). This reveals a critical limitation of
the state-of-the-art models in maintaining consistency.


In particular, we motivate our definition of consistency based on classical deductive logic
\cite{tarski1994introduction} that defines a consistent theory as one that does not entail a contradiction. Correspondingly, we define
consistency, in the context of VQA, as being able to answer questions posed from different semantic perspectives about
a certain fact without any contradiction. In addition, consistent Question-Answer (QA) pairs can be derived based on simple notions of logic
or by commonsense reasoning. For instance, say an image contains a ``large building''. Logic-based QA pairs
can be ``is the building small? no" and ``what size is the building? large". On the other hand, if an image contains ``vegetarian pizza'', commonsense-based QA pairs can be ``is it a vegetarian pizza? yes'' and ``is there
pepperoni on the pizza? no'', which requires commonsense knowledge that ``pepperoni'' is not vegetarian.

\begin{figure}[t]
\centering
\includegraphics[height=61px]{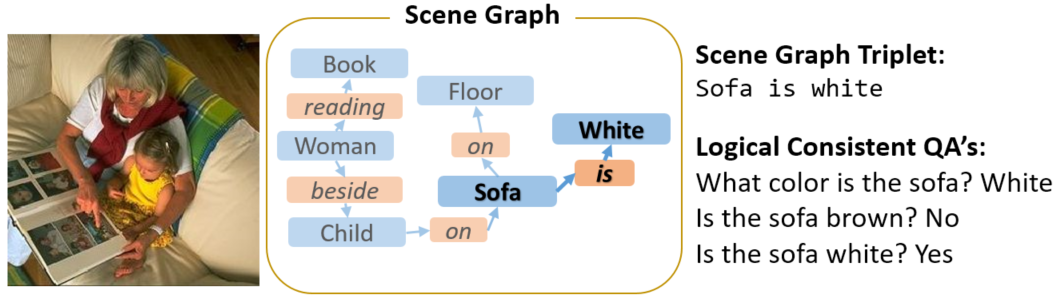}
\vspace{2mm}
\caption{Qualitative examples from our ConVQA dataset derived based on logic.}
\label{fig:Logicdatasetexamples}
\vspace{-2mm}
\end{figure}

\begin{figure}[t]
\centering
\includegraphics[height=125px]{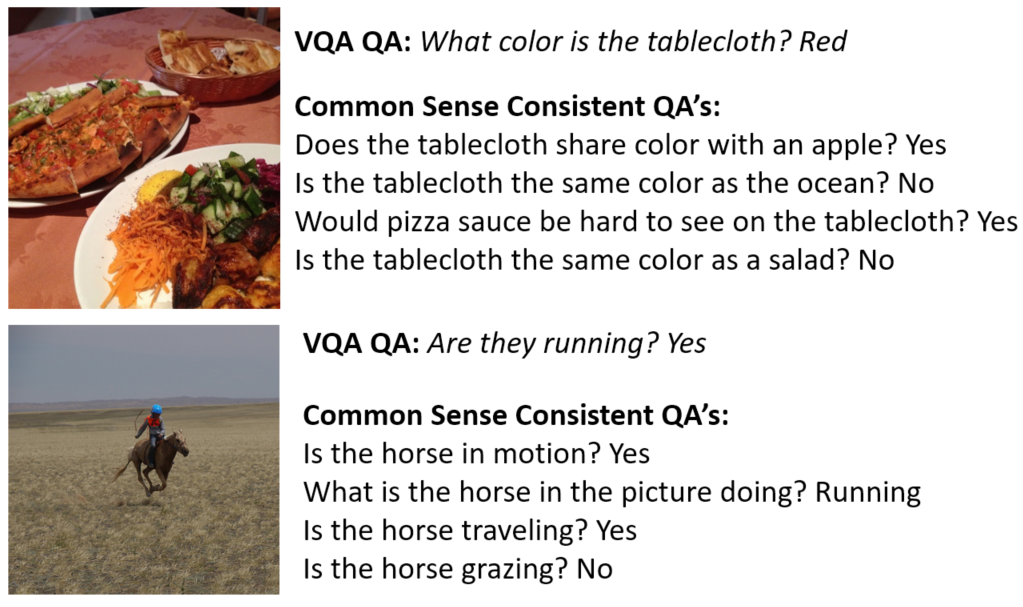}
\vspace{2mm}
\caption{Qualitative examples from our human-annotated commonsense-based ConVQA Dataset.}
\label{fig:CSdatasetexamples}
\vspace{-2mm}
\end{figure}

While attempts have been made to construct logic-based consistent VQA datasets \cite{hudson2019gqa}, they still fall short on commonsense-based consistency. To
this end, our ConVQA Dataset consists of two subsets: 1) a challenging human-annotated set comprised of
commonsense-based consistent QA's (shown in Figure \ref{fig:CSdatasetexamples}), and 2) an automatically generated logic-based consistent QA dataset (shown in Figure \ref{fig:Logicdatasetexamples}). 


To improve the consistency of VQA models, we propose a Consistency Teacher Module (CTM), which consists of a Question Generator that synthesizes entailed (or similar-intent)
questions given a seed QA pair and a Consistency Checker that examines whether answers to those similar-intent questions are consistent. For training a consistent VQA model, our CTM acts as a consistency-based data augmentation scheme that trains a VQA model with consistent answers to entailed questions. We demonstrate that our approach
improves the performance of a baseline VQA model on our ConVQA testing sets in terms of both accuracy and consistency. Our datasets and models will be available at https://bit.ly/32exlM7. 



\begin{figure*}[h!]
\centering
\includegraphics[height=70px]{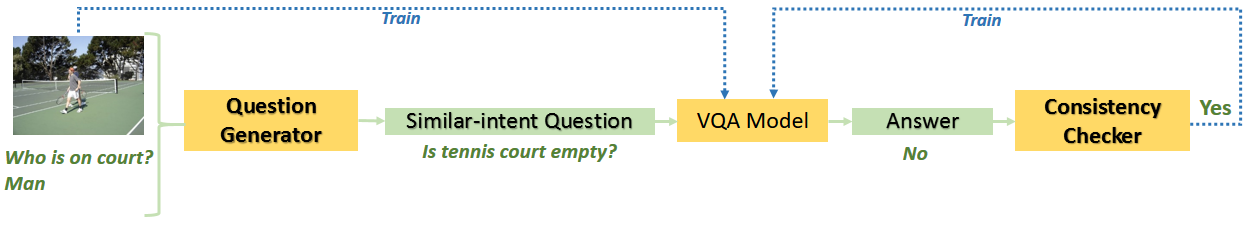}
\vspace{1.8mm}
\caption{Block diagram of the proposed CTM including a Question Generator that synthesizes questions with similar intent and a Consistency Checker that classifies QA pairs as consistent, unrelated, or contradictory. CTM finetunes VQA models via reinforcement learning with the answer consistency as the reward to encourage VQA models to answer rephrased questions more consistently. The examples shown are from a real run.}
\label{fig:QA2Q}
\end{figure*}

\vspace{\sectionReduceTop}
\section{Related Work}
\vspace{\sectionReduceBot}
\label{sec:related_work}

Checking for consistency can be considered as an interrogative Turing Test \cite{radziwill2017evaluating} for linguistic robustness \cite{stede1992search}, \cite{li2017robust}.
Works such as \citet{xu2018fooling} explore the robustness of VQA with respect to image variations,
whereas works such as \citet{ray2016question} and \citet{mahendru2017promise} focus on the understanding of the premise of a question instead of relying on dataset biases \cite{agrawal2017c} \cite{goyal2017making} or linguistic biases \cite{ramakrishnan2018overcoming}.


Recently, the research community has shown great interest in evaluating VQA for
consistency and plausibility. GQA \cite{hudson2019gqa} is established as a scene-graph
based QA dataset. Their questions, similar to \citet{johnson2017clevr},
require multiple hops of reasoning, and are not validated or annotated by humans.
Our ConVQA differs from GQA in the following two aspects. First, we provide a human-validated test
set of the automatically generated logic-based consistent QA's for a more accurate performance evaluation. Second, we collect human-annotated QA pairs based on common-sense in
addition to the logic-based QA's. The most relevant work to ours is
\citet{shah2019cycle}. However, they focus strictly on question paraphrases that maintains the same answers as the source question. 
We, however, focus on generating questions which can have different answers, but are about the same visual fact, which greatly increases the diversity of the resulting QA pairs.
To the best of our knowledge,
the proposed ConVQA dataset is the first consistent QA dataset that contains human-annotated consistent QA's based on common-sense. 

Other works have also looked into question generation \cite{zhang2016automatic}, \cite{mostafazadeh2016generating} for training better VQA models. In \citet{misra2017learning}, QA pairs are obtained from an oracle in a simulated environment. In contrast, our CTM-based training operates on real images and uses a learned consistency measure to train the VQA module with consistent QA's.


\vspace{\sectionReduceTop}
\section{ConVQA Datasets}
\vspace{\sectionReduceBot}
\label{sec:datasets}
The consistent QA pairs in our ConVQA are generated automatically based on simple notions of logical consistency or are human-annotated using commonsense reasoning.

\xhdr{Logic-based Consistent QA. (L-ConVQA)} 
Consider the Visual Genome \cite{krishna2017visual} scene graph in Figure \ref{fig:Logicdatasetexamples} consisting of objects, attributes, and their relationships.  We consider each triplet to encode a single `visual fact', for instance, that the sofa is white. 
We employ slot-filler NLP techniques to generate a set of QA pairs for each triplet (object-relation-subject)
in the scene graph. Currently, we focus on attribute (e.g., color, size), existential (e.g.,
is there) and relational (e.g., sofa on floor) consistency. We leverage Wordnet~\cite{miller1995wordnet} and a manually generated list of antonyms (e.g.,
white vs. black) and hypernyms (e.g., white $\rightarrow$ color) to generate these QA pairs. For example, for the
attribute ``white'' of an object ``cup'', we generate QA pairs such as ``is the cup white? yes'', ``is the cup black?
no'' and ``what color is cup? white''. We also filter objects and relationships by frequency and saliency (e.g., based
on bounding boxes) to avoid non-salient and infrequent objects or noisy relationships. We have a total of 880,141 QA
pairs in 255,910 sets on 70,292 images. We split the data into a training set with 47,999 images, a validation set with 9,993 images, and a test set with 12,300 images.  Notably, we create a smaller clean test set (12,325 QA pairs on 725 images) using Amazon
Mechanical Turk (AMT) where three independent workers were asked to remove incorrect or unnatural QA's.

\vspace{-1pt}
\xhdr{Commonsense-based Consistent QA.}
While the logic-based consistent QA set provides a first step into large-scale examination of VQA consistency, the generated questions require limited reasoning and commonsense and are, therefore, frequently simpler than human-annotated ones.
Hence, we collect more challenging QA pairs based on commonsense (\textbf{CS-ConVQA}) by asking AMT workers to write intelligent rephrases
of QA pairs sampled from the VQA2.0 \cite{goyal2017making} validation dataset. AMT workers were instructed to avoid simple word paraphrases and instead to write rephrases that require commonsense reasoning in order to answer the question consistently. We collect approximately 3.5 consistent QA pairs per image for 6439 images. After filtering images that overlap with the training set of the L-ConVQA subset, we split this data into a training set (1568 images), a validation set (450 images), and a test set (1590 images). 
\vspace{-2pt}

\vspace{\sectionReduceTop}
\section{Approach}
\vspace{\sectionReduceBot}
\label{sec:approach}

To improve VQA consistency, we propose training a VQA model using a Consistency Teacher Module (CTM) that generates entailed questions and performs a consistency-based data augmentation. More specifically, CTM consists of two trainable components -- an entailed question generator and a consistency checker. 

\xhdr{Entailed Question Generator.} For a given a source question-answer pair, we define entailed questions as those for which the answer should be obvious given the source QA pair. For example, given the source QA pair ``Who is on court? Man'', an appropriate entailed question might be ``Is the tennis court empty?''. We train a question generation model that given representations of the image and the source QA pair, generates a new question. Specifically, our question generator concatenates the deep features of an image 
(extracted
using a ResNet152 \cite{he2016deep} network) 
and a QA pair (extracted using a 1-layer LSTM \cite{hochreiter1997long}) to represent a visual fact. These features are fed
into another LSTM model to generate a similar-intent question. We train this module on the automatically generated Logical L-ConVQA train set. We also include some closely related (according to averaged Word2Vec \cite{mikolov2013distributed} distance) Visual Genome \cite{krishna2017visual} QA pairs in the training of the question generator to add some diversity to the generated questions.

\xhdr{Consistency Checker.} Once the VQA model produces an answer for the generated entailed question, it may or may not be consistent with the source question. To evaluate this and provide feedback to the VQA model, we train a consistency checker that processes the image and both the source and entailed QA pairs. Similar to the question generator architecture, the consistency checker takes in deep features of the image and two QA pairs and classifies the QA pairs as consistent, inconsistent or unrelated. This model is trained using the automatically generated L-ConVQA train set alone. Inconsistent examples in the L-ConVQA set are made using simple techniques such as flipping yes/no answers and
replacing entities in the scene graph triplets.

\xhdr{Consistency Teacher Module (CTM).} Putting these two components together, we can train a VQA model based on its consistency on generated entailed questions. Figure \ref{fig:QA2Q} shows our pipeline. During training, for each source VQA QA pair (``Who is on court? Man''), we generate an entailed question (``Is tennis court empty?'') and produce the VQA model's answer (``No''). We then run the consistency checker to determine if the generated answer is consistent with the source QA (in this case, ``Who is on court? Man''). If the answer is consistent (and VQA confidence $>0.7$), we treat it as the ground truth for the entailed question and update the VQA model as if this example were part of the original dataset. Likewise if it is deemed inconsistent, or if the question is deemed unrelated, it is unclear what the correct answer should be, so we do not update the model.

\vspace{\sectionReduceTop}
\section{Experiments}
\vspace{\sectionReduceBot}
\label{sec:experiments_results}
\begin{figure}[t]
\centering
\includegraphics[height=170px]{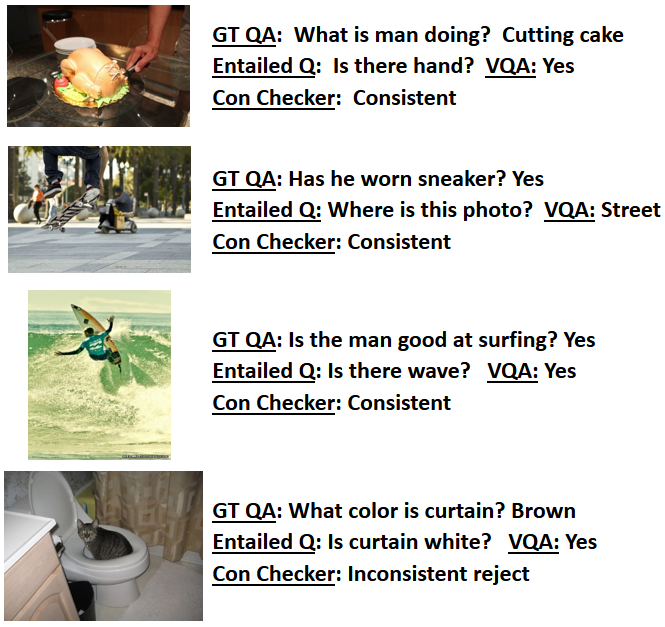}
\caption{Qualitative examples of entailed question generation and consistency checking for training the VQA.}
\label{fig:qualExamplesCTM}
\vspace{-2mm}
\end{figure}

\begin{table*}[]
\caption{Performance comparison of baseline VQA trained on VQA2.0, baseline VQA finetuned on ConVQA, and VQA trained using our CTM. \textbf{L-ConVQA} is the human-cleaned Logical Consistent QA dataset, \textbf{CS-ConVQA} is the human annotated Common-sense Consistency Dataset and \textbf{VG} is Visual Genome. CTM-based training produces the best results in terms of overall accuracy and consistency. \textbf{DATA} denotes the data used to fine-tune VQA or seed the CTM question generator.}
\label{Table:ExpResults}
\centering
\renewcommand*{\arraystretch}{0.99}
\setlength{\tabcolsep}{3pt}
\vspace{3pt}
\resizebox{\textwidth}{!}{
\begin{tabular}{clc@{\hspace{15pt}}cccc@{\hspace{10pt}}ccccc}
\toprule
                      &          & \small{\textbf{DATA}} & \multicolumn{3}{c}{\small\textbf{L-ConVQA}}   &       & \multicolumn{5}{c}{\small\textbf{CS-ConVQA}}                                                                            \\\cline{4-6}\cline{8-12}
                     &          & & \footnotesize\textbf{Perf Con}    & \footnotesize\textbf{Avg Con}     &\footnotesize\textbf{Top1} &     & \footnotesize\textbf{Perf Con}    & \footnotesize\textbf{Avg Con}     & \footnotesize\textbf{Top1}        & \footnotesize\textbf{Yes/No} & \footnotesize\textbf{Num}    \\
                                \midrule
                               
\footnotesize{a)} & \small{VQA}    & \small{VQA2.0}              & 36.25                & 71.36                & 70.34       &         & 26.13                & 59.61                & 60.03                & 65.49                               & 31.39           \\
\cline{1-12}
\footnotesize{b)} &\small{FineTune} & \small{CS-ConVQA}                     & 34.54                & 70.39                & 69.48    &            & 26.39                & 59.65                & 60.07                & 65.80                               & 35.92           \\    
\footnotesize{c)} &\small{FineTune} & \small{L/CS-ConVQA}                 & \textbf{54.68}                & \textbf{83.42}                & \textbf{83.16}     &           & 24.70                 & 59.30                 & 59.60                 & 65.14                               & 33.33           \\
\footnotesize{d)} &\small{\textbf{+CTM}} &  \small{L/CS-ConVQA}            & 54.6                & 83.23                & 82.79       &         & 25.94                & \textbf{60.39}                & \textbf{60.78}                & \textbf{66.63}                               & \textbf{36.89}  \\
\footnotesize{e)} &\small{FineTune} & \small{L/CS-ConVQA,VG}   & 36.40 & 71.60 & 70.94 & & 25.22 & 59.19 & 59.56 & 65.30 & 31.39            \\
\footnotesize{f)} &\small{\textbf{+CTMvg}} & \small{L/CS-ConVQA,VG}    & 51.41                & 81.66                & 81.37    &            & \textbf{27.49}        & 59.75                & 60.15                & 66.41                               & 34.95 \\         
\bottomrule
\end{tabular}}
\end{table*}

\begin{figure}[t]
\centering
\includegraphics[height=225px]{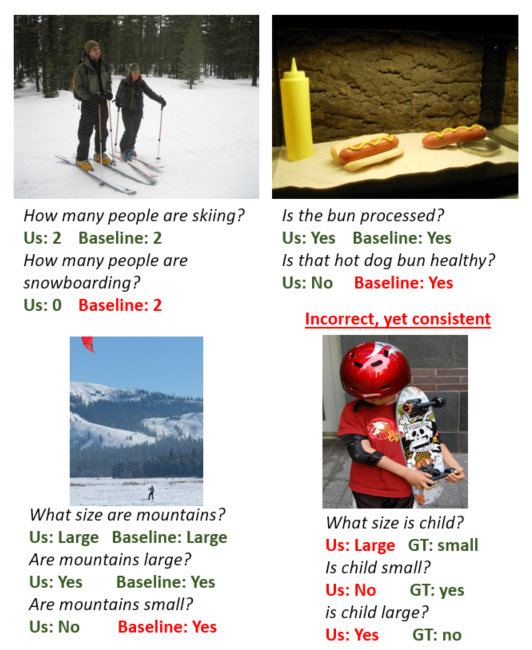}
\vspace{2mm}
\caption{Qualitative examples of our improved VQA consistency by CTM training compared to baseline bottom-up VQA. Green is correct, Red is wrong. GT is Ground Truth.}
\label{fig:qualExamples}
\vspace{-2mm}
\end{figure}

To evaluate our approach, we apply the Consistency Teacher Module (CTM) module to a state-of-the-art VQA model trained on VQAv2 and evaluate performance on the ConVQA datasets. We describe training procedures, metrics, and baselines in this section.

\xhdr{Consistency Teacher Module Training.} We train the components of CTM -- Entailed Question Generation and Consistency Checker -- using only the synthetic L-ConVQA train set (referred as the standard \textbf{CTM}) or a mix of Visual Genome and L-ConVQA train (referred as \textbf{CTMvg}) and keep them frozen when fine-tuning the VQA model. When we train the VQA, the sets used to finetune the VQA or seed the CTM come from splits not seen during training of the CTM - val split of L-ConVQA and train split of CS-ConVQA.

Qualitative examples of our Entailed Question Generator trained only on L-ConVQA are shown in Figure \ref{fig:qualExamplesCTM}. Despite only being trained on the automatically-generated L-ConVQA data, it generates reasonably well-entailed questions on human-annotated questions.

Our Consistency Checker has a high accuracy of classification on the L-ConVQA test set ($90\%$). However, when tested with a mix of commonsense-based CS-ConVQA, the accuracy drops to $64\%$ (chance is $33\%$ for $3$ classes). 
We find that precision is important when training the VQA using the pre-trained Consistency Checker. Hence, we use the classifier at above $90\%$ confidence threshold, where the precision is $70.38\%$.

\xhdr{Evaluation Metrics for ConVQA.} We report three metrics for ConVQA -- capturing notions of consistency and performance. 

\begin{compactenum}[--]
\item \textbf{Perfect-Consistency (Perf-Con).} A model is perfectly consistent for a question set if it answers all questions in the set correctly. We report the percentage of such sets as Perf-Con.
\item \textbf{Average Consistency (Avg-Con).} We also report the average accuracy within a consistent question set over the entire dataset as Avg-Con.
\item \textbf{Accuracy (top-1).} Finally, we report the top-1 accuracy over all questions in the dataset. 
\end{compactenum}

\xhdr{Baselines.} We compare to a number of baseline models to put our CTM results into context:
\begin{compactenum}[--]
\item \textbf{VQA.} We take the bottom-up top-down VQA model \cite{anderson2018bottom} as our base model for these experiments. To evaluate consistency in existing models, we present results on ConVQA of this model pretrained on VQA2.0. 

\item \textbf{Finetuned models:} We present results for models finetuned on ConVQA and Visual Genome -- \textbf{Finetune CS-ConVQA} finetuned on the commonsense ConVQA dataset, \textbf{Finetune L/CS-ConVQA} on both logical and commonsense ConVQA, and \textbf{Finetune L/CS-ConVQA,VG} extending to Visual Genome questions. 
\end{compactenum}
\noindent When we apply our CTM model to the finetuned baselines above, we seed the question generator with the associated dataset.

\section{Results and Analysis}

Table \ref{Table:ExpResults} shows quantitative results on our L-ConVQA and CS-ConVQA datasets. We make a number of observations below.

\xhdr{The state-of-the-art VQA has low consistency.} The baseline VQA system \textbf{(row a)} retains similarly high top-1 accuracy on the ConVQA splits (63.58\% on VQAv2 vs 70.34\% / 60.03\% on L-ConVQA / CS-ConVQA); however, it achieves only 26.13\% perfect consistency on the human generated CS-ConVQA questions. 

\xhdr{Finetuning is an effective strategy for the synthetic L-ConVQA split.} Finetuning on L-ConVQA train results in 18.43\% gains in perfect consistency on L-ConVQA test (\textbf{row c vs a}). This is unsurprising given the templated questions and simple concepts in L-ConVQA; however, perfect consistency is low in absolute terms at 54.68\%. 

\xhdr{Finetuning does not lead to significant gains in consistency for human-generated questions.} Finetuning the VQA model on CS-ConVQA \textbf{(row b)} leads to an improvement in consistency of only 0.26\%.  Likewise, adding L-ConVQA \textbf{(row c)} and extra Visual Genome questions \textbf{(row e)} actually reduces consistency.

\xhdr{CTM-based training preserves or improves consistency when leveraging additional data.} When we apply CTM to the Finetuned L/CS-ConVQA model, we improve CS-ConVQA perfect consistency by 1.24\% \textbf{(row d vs c)} while modestly improving other metrics. Extending to Visual Genome questions, the CTM augmented model improves perfect consistency in CS-ConVQA by 2.27\% over the finetuned model \textbf{(row f vs e)}. Interestingly, the CTM modules were never trained with the human-annotated CS-ConVQA questions and yet lead to this improvement on CS-ConVQA by acting as an intelligent data augmenter/regularizer.


\vspace{-3pt}
\section{Conclusion and Discussion}
\vspace{-3pt}

In this paper, we introduced a ConVQA dataset consisting of logic-based and commonsense-based consistent QA pairs about
visual facts in an image. We also proposed a Consistency Teacher Module that acts as a consistency-based data augmenter to teach VQA models to answer consistently. As future work, we plan to look into
improving our automatically generated consistent QA pairs using external knowledge-bases. 

\vspace{-1mm}
\ifaclfinal
\section*{Acknowledgements}
\vspace{-1mm}
We thank Yi Yao, Xiao Lin, Anirban Roy, Avi Ziskind and the anonymous reviewers for their helpful comments. This research was developed under DARPA XAI. The views, opinions and/or findings expressed are those of the authors' and should not be interpreted as representing the official views/policies of the DoD / U.S. Govt.
\fi

\bibliography{emnlp-ijcnlp-2019}
\bibliographystyle{acl_natbib}

\end{document}